%% file: acl_latex.tex
\documentclass[11pt]{article}

\usepackage[final]{acl}

\usepackage{times}
\usepackage{latexsym}
\usepackage[T1]{fontenc}
\usepackage[utf8]{inputenc}
\usepackage{microtype}
\usepackage{inconsolata}
\usepackage{graphicx}

\usepackage{amsmath,amsfonts,bm}

\usepackage{booktabs}
\usepackage{multirow}
\usepackage{array}

\usepackage[table]{xcolor}

\def\method{\textsc{Sayre}}

\title{Enhancing Large Multimodal Models in Key Information Extraction \\via Scene-Aware Document Synthesis}

\author{
Zhipeng Xu$^{\spadesuit}$, Zulong Chen$^{\spadesuit}$, Qing Liu$^{\spadesuit}$, Junhao Ji$^{\spadesuit}$, \\
\textbf{Jinxin Hu$^{\spadesuit}$, Yipeng Yu$^{\diamondsuit}$, Jianqiang Wan$^{\heartsuit}$, Jun Tang$^{\heartsuit}$ and Zhao Li$^{\clubsuit}$} \\
$^{\spadesuit}$Alibaba Group \ \
$^{\heartsuit}$Qwen Team, Alibaba Group \\
$^{\diamondsuit}$Taobao and Tmall Group, Alibaba \ \
$^{\clubsuit}$Zhejiang University
}

\begin{document}
\maketitle

\input{sections/abstract}
\input{sections/intro}
\input{sections/related_work}
\input{sections/methodology}

\input{sections/experiment}
\input{sections/result}
\input{sections/conclusion}

\bibliography{custom}

\end{document}

%% file: sections/abstract.tex
\begin{abstract}
Key Information Extraction (KIE) converts visually rich documents into structured data, but practical deployment remains challenging: strong performance often relies on costly on-server Large Multimodal Models (LMMs), while compact locally deployable models lack sufficient KIE supervision.
We present \method{}, a scene-aware document synthesis framework for generating scalable KIE training data without hand-crafted template design. Given a few exemplar documents, \method{} captures category-specific content patterns and layout conventions to synthesize document--schema--annotation triples. It further introduces error-driven generation, which expands real-world failure cases into hard training examples while preserving their structural patterns.
Experiments on constrained- and open-category KIE show that \method{} consistently improves Qwen3-VL backbones and achieves the strongest overall performance among on-device LMMs. Data scaling experiments show an overall upward trend as more synthesized data is introduced, especially for smaller models and open-category extraction. Error analysis further shows that synthesized training reduces field-level errors by improving schema-aware extraction over dense tables, business identifiers, and contract clauses. These results establish scene-aware synthesis as an effective data-centric approach for improving practical multimodal KIE.
\end{abstract}

%% file: sections/intro.tex
\section{Introduction}

Key Information Extraction (KIE) is a fundamental task in document intelligence that converts visually rich documents into structured, actionable data for downstream applications such as invoice processing, contract management, and financial document analysis~\cite{rombach2025deep,huang2019icdar,palm2017cloudscan,stanislawek2021kleister}. Recent Large Multimodal Models (LMMs) have demonstrated strong potential for end-to-end KIE by jointly modeling textual content, visual appearance, and layout structure~\cite{cao2023genkie,wang2024docllm,bhattacharyya2025information,lu2025bounding}. However, state-of-the-art performance often depends on large on-server models, whose high inference costs, latency, and data-transfer requirements hinder deployment in many enterprise settings~\cite{guo2018cloud,kang2017neurosurgeon}. This limitation has stimulated growing interest in smaller LMMs that can be deployed locally. Yet, their extraction capabilities remain insufficient, particularly for documents involving diverse layouts, domain-specific schemas, and long-tail field configurations~\cite{ji2026unikie,shen2026ocr,zmigrod2024value,jiang2025relayout}.

A key bottleneck is the scarcity of scalable training data for KIE~\cite{rombach2025deep,skalicky2022business,simsa2023docile,dua2025flexdoc}. Collecting and annotating real enterprise documents at scale is challenging, as KIE supervision requires not only field values but also extraction schemas and field-level correspondences. Document synthesis offers a promising alternative; however, existing approaches typically rely on manually designed templates or simple content replacement strategies~\cite{bensch2021kie,simsa2023docile,dua2025flexdoc,zmigrod2024value}. These methods are costly to scale across document categories and often fail to preserve the category-specific content patterns and layout conventions of real documents~\cite{bensch2021kie,simsa2023docile,dua2025flexdoc}. Moreover, the synthesized data may lack sufficient diversity and provide limited coverage of the challenging cases encountered in practical KIE systems~\cite{dua2025flexdoc,zmigrod2024value,ji2026unikie,shen2026ocr}.

To address this issue, we propose \method{}, a scene-aware document synthesis framework that combines general data generation with error-driven data generation. For general generation, \method{} takes only a few exemplar documents from a target document category, perceives their content patterns and layout conventions, and uses these category-level cues to generate new documents together with their corresponding extraction schemas and structured annotations. This enables scalable data expansion without hand-crafted templates. To further cover challenging scenarios that general synthesis may miss, we introduce error-driven generation based on failure cases collected from model testing and production systems. It retains the challenging page structures and field organizations of these cases while rewriting their content and updating the corresponding annotations, thereby expanding a limited set of real-world failure patterns into a larger collection of hard training examples.

Extensive experiments across diverse KIE categories and extraction settings demonstrate the effectiveness of \method{}. Augmenting training data with the synthesized examples consistently improves LMM performance across different model scales, with particularly substantial gains for smaller models. These results indicate that the generated samples provide useful supervision for improving KIE under varied document categories and extraction settings. Further experiments demonstrate the scalability of \method{}: model performance shows an overall upward trend as the amount of synthesized training data increases, suggesting that additional generated data can yield sustained benefits beyond small-scale augmentation. This scaling is achieved without relying on hand-crafted templates. Taken together, these results establish \method{} as an effective and scalable data-centric approach for advancing practical KIE.

%% file: sections/related_work.tex
\section{Related Work}
Key Information Extraction (KIE) aims to automatically identify and extract structured fields from visual documents, such as invoices, receipts, and forms~\citep{grishman1997information,simon2005viper,aumann2006visual}. It serves as a fundamental component for a wide range of enterprise downstream applications, and has attracted substantial attention from both researchers and practitioners~\citep{chiticariu2010enterprise,cui2021document,skalicky2022business,rombach2025deep}. Early approaches formulate the KIE task in visual documents as a two-stage pipeline~\citep{zhang2020trie,abdallah2024survey}. They typically employ Optical Character Recognition (OCR) tools to extract textual content from document images, and then cast KIE as a sequence-labeling task over the recognized tokens~\citep{palm2017cloudscan,hwang2021cost,zhang2023reading}. While effective, these methods primarily rely on the semantic content of extracted text, overlooking crucial visual layout cues that are essential for accurate field extraction~\citep{li2021structext,peng2022ernie,wang2024docllm}.

To address this limitation, some researchers explore incorporating document layout features to capture relationships among textual elements~\citep{katti2018chargrid,xu2020layoutlm,li2021structext,hong2022bros,wang2022lilt}. These approaches still rely on OCR outputs, but jointly encode textual content with their two-dimensional positional information, enabling models to reason over both semantic and spatial structures within documents~\citep{sun2021spatial,xu2021layoutxlm,li2021selfdoc,shen2022vila}. Despite their effectiveness, such layout-aware language models remain heavily dependent on OCR accuracy, making them vulnerable~\citep{bhattacharyya2025information,barboule2025survey,ding2026deep}. Further efforts attempt to mitigate this issue by directly incorporating raw visual signals from document images into the model~\citep{appalaraju2021docformer,huang2022layoutlmv3,shen2022vila}. Nevertheless, since they still rely on OCR text as the primary input modality, their robustness remains limited when recognition errors occur~\citep{agarwal2025fs,shim2025revise,shen2026ocr,liu2026see}.

More recent efforts explore the use of Large Multimodal Models (LMMs) to perform KIE on visual documents~\citep{cao2023genkie,hu2024mplug,bai2025qwen3,liu2026textmonkey}. These methods directly take document images as input to jointly model textual, visual, and layout features, demonstrating superior effectiveness~\citep{xie2024wukong,zhang2024document,ke2025large}. Prior work enhances the perception of LMMs by adopting document-oriented pretraining~\citep{li2022dit,mao2024visually,yu2025minicpm,bai2025qwen2}, allowing LLMs to capture layout cues in the document~\citep{li2024monkey,liu2024focus,zhang2025dockylin,xie2024wukong}. However, effective extraction requires models to reason over documents, identify relevant entities, and understand their relationships~\citep {zhang2024modeling}. Recent advances explore leveraging the reasoning capabilities of LMMs and external toolkits to deepen document understanding~\citep{han2025mdocagent,wang2025vrag,xiong2026lang2act}. Nevertheless, these approaches typically rely on scaling model size, which introduces additional computational overhead and latency.

%% file: sections/methodology.tex
\section{Scene-Aware Document Synthesis for Key Information Extraction}
We introduce \method{}, a manual-template-free framework for KIE data synthesis. Section~\ref{general} describes exemplar-guided generation, while Section~\ref{error} transforms real-world failure cases into challenging training samples. We further simulate practical document conditions by adding visual degradations to the synthesized documents.

\subsection{General Data Generation.} 
\label{general}
We depart from prior methods that rely on hand-crafted templates to replicate documents, and instead introduce a scene-aware document data synthesis framework.
Given a few exemplar documents from a specific category $c$ collected from an enterprise document-processing system, our framework can synthesize diverse document instances of the same category, along with their corresponding query schemas $\mathcal{S}$ and structured outputs $y$, which can be expressed as: 
\begin{equation}
\small
(x, \mathcal{S}, y) \sim \mathcal{G}(\{x_i\}_{i=1}^n, \mathcal{I}_c,  \mathcal{P}), \quad x_i \in \mathcal{X}_c \ \forall i,
\end{equation}
where $\mathcal{G}$ denotes our synthesis framework, $n$ denotes the number of exemplar documents, $\mathcal{X}_c$ represents the set of documents belonging to category $c$, $\mathcal{I}_c$ denotes the document category identifiers and $\mathcal{P}$ is the persona card, aiming to increase the diversity of the generated content. 
To capture the complex semantics and visual features of document images, $\mathcal{G}$ is instantiated as a multi-agent system in which multiple agents jointly perform perception and generation to generate new documents. Specifically, we first prompt the agent $V_{t}$ to generate a new topic $t$ based on the identifier $\mathcal{I}_c$ and the persona card $\mathcal{P}$, which can be expressed as:
\begin{equation}
\small
t = V_{t}(\mathcal{I}_c, \mathcal{P}).
\end{equation}
Then we use the VLM-based  content perception agent $V_\mathcal{C}$ and the layout perception agent $V_\mathcal{L}$ to generate the description about the exemplar documents:
\begin{equation}
\small
\mathcal{C} = V_\mathcal{C}(\{x_i\}_{i=1}^n);\quad
\mathcal{L} = V_\mathcal{L}(\{x_i\}_{i=1}^n),
\end{equation}
where content description $\mathcal{C}$ summarizes the key semantic elements of the document category, including commonly appearing fields, their semantic meanings, and their potential dependencies, while layout description $\mathcal{L}$ captures the global organization of the page, such as page size, hierarchical structure, and the spatial distribution of elements. We then employ the content generation agent $V_\mathcal{J}$ to produce the structured outputs $y$ conditioned on the content description $\mathcal{C}$ and the newly sampled topic $t$, and the corresponding query schema $\mathcal{S}$ is then derived from $y$ by removing the value fields while retaining the structure:
\begin{equation}
\small
y = V_\mathcal{J}(\mathcal{C}, t, \mathcal{P});\quad \mathcal{S} = \mathrm{Schema}(y).
\end{equation}
Subsequently, we utilize the document generation agent $V_\mathcal{D}$ to synthesize the document by integrating the structured outputs $y$ with the layout description $\mathcal{L}$. Specifically, the agent generates HTML code $\mathcal{H}_x$ that is further rendered into the final document image $x$:
\begin{equation}
\small
\mathcal{H}_x = V_\mathcal{D}(y, \mathcal{L});\quad x = \mathrm{Render}(\mathcal{H}_x).
\end{equation}
For domain-specific scenarios, we further perform extraction over $y$ using a predefined schema $\mathcal{S}'$, resulting in scenario-specific structured outputs $y'$, thereby improving the relevance of the synthesized data to the target application, which can be expressed as:
\begin{equation}
\small
y'= V_\mathcal{A}(y, \mathcal{S}'),
\end{equation}
where $V_\mathcal{A}$ denotes a schema-guided extraction agent that maps the structured outputs $y$ to the target schema $\mathcal{S}'$ by selecting and reorganizing the relevant fields, ensuring consistent field alignment and semantic coherence. This additional extraction step enables our framework to flexibly adapt the synthesized data to diverse downstream scenarios by focusing on task-relevant fields, effectively bridging the gap between general-purpose document generation and application-specific supervision.

\subsection{Error-driven Data Generation}
\label{error}
While the general data generation produces massive training data, it primarily focuses on improving the general document extraction capability of the model and does not explicitly capture real-world failure modes. To address this limitation, we curate a corpus of real-world failure cases and introduce an error-driven data generation strategy grounded in failures observed in real-world scenarios.

We construct a corpus of real-world failure cases through two sources. First, we train the model on general synthetic data, deploy it for testing, and collect the resulting failure cases. Second, we curate failure cases from our in-house production document processing system, which aggregates errors arising from multiple processing pipelines, including OCR-based extraction methods and end-to-end approaches. All collected failure cases are manually annotated to ensure high-quality supervision for subsequent error-driven data generation.

Then we follow~\citet{poznanski2025olmocr} to prompt an agent $V_{\mathcal{E}}$ to transform each collected failure case in the corpus $\mathcal{X}_{f}$ into an HTML-based template, which can be expressed as:
\begin{equation}
\small
\mathcal{H}_{x'} = V_{\mathcal{E}}(x'), \quad x' \in \mathcal{X}_{f},
\end{equation}
where $\mathcal{H}_{x'}$ represents the final refined HTML template.
We apply a set of parsing rules to extract all textual content $\mathcal{T}$ from the HTML code, and align these extracted text blocks with the values in the labels $y'$ to construct a mapping $\mathcal{M}$. This process can be defined as:
\begin{equation}
\small
\mathcal{T} = \mathrm{Parse}(\mathcal{H}_{x'});\quad \mathcal{M} = \mathrm{Match}(\mathcal{T},y')
\end{equation}
The extracted text blocks $\mathcal{T}$ are then fed to an LLM, which rewrites them into semantically similar yet fully distinct content $\tilde{\mathcal{T}}$ for de-identification:
\begin{equation}
\small
\tilde{\mathcal{T}} = {\mathrm{LLM}}(\mathcal{T}).
\end{equation}
Based on the mapping $\mathcal{M}$ and the rewritten text $\tilde{\mathcal{T}}$, the original labels $y'$ are updated to $\tilde{y}'$ to maintain alignment with the new content. This update process can be formulated as:
\begin{equation}
\small
\tilde{y}' = \mathrm{Update}(y', \mathcal{M}, \tilde{\mathcal{T}});\quad \mathcal{S'} = \mathrm{Schema}(\tilde{y}')
\end{equation}
Finally, the rewritten text $\tilde{\mathcal{T}}$ is reinserted into the HTML template, replacing the original content $\mathcal{T}$ to yield a new synthetic HTML template $\tilde{\mathcal{H}}_{x'}$, which is then rendered into the final synthetic document instance $\tilde{x}'$:
\begin{equation}
\small
\tilde{\mathcal{H}}_{x'} = \mathrm{Replace}({\mathcal{H}}_{x'}, \mathcal{T}, \tilde{\mathcal{T}}); \quad\tilde{x}' = \mathrm{Render}(\tilde{\mathcal{H}}_{x'})
\end{equation}
For samples in which certain fields in $y'$ cannot be reliably mapped, we adopt a multi-model voting strategy for re-annotation. Specifically, we leverage multiple advanced LMMs to re-extract the corresponding values from the document image and select the most frequently predicted answer as the final annotation.

%% file: sections/experiment.tex
\section{Experimental Methodology}
In this section, we describe the baselines, benchmarks, evaluation metrics, and implementation details of our experiments.

\textbf{Baselines.}
We compare the model trained via \method{} with representative on-device LMMs as baselines, including MiniCPM-V4.5-8B~\citep{yu2025minicpm}, InternVL3.5-8B~\citep{wang2025internvl3}, Qwen3-VL-2B and Qwen3-VL-4B~\citep{bai2025qwen3}, Ministral-3-8B~\citep{liu2026ministral}, MiMo-VL-7B-RL~\citep{team2025mimo}, and GLM-4.1V-9B~\citep{team2025glm}.
In addition, several advanced on-server LMMs are included as upper-bound baselines, including GPT-4o, GPT-5, Claude-Sonnet-4.5, Qwen-VL-Max, Qwen3-VL-Plus, and Gemini-3-Pro.

\textbf{Benchmarks.}
We evaluate the model trained via \method{} on UniKIE~\citep{ji2026unikie}, a benchmark for universal KIE that spans diverse document types and heterogeneous field definitions. 

\textbf{Evaluation Metrics.}
We follow~\cite{yang2025cc} to evaluate the KIE performance of the \method{} models and the baseline LMMs using the field-level F1 score~\citep{hwang2019post,xu2020layoutlm}.
A predicted field is considered correct only if the extracted value exactly matches the ground-truth annotation after normalization.

\textbf{Implementation Details.}
\input{tables/overall_merged}
We employ Qwen-VL-Max to perceive document layout and semantic content in the general data generation, and use Qwen3-Max to generate the textual content and corresponding HTML code for new documents. We sample 1M elite personas from the persona hub proposed by~\cite{ge2024scaling} to improve the diversity of generated documents. In the error-driven data synthesis pipeline, Qwen3-VL-Plus is used to generate document templates. Together, the two synthesis pipelines generate 1M data instances, which are further augmented in Blender with realistic optical noise to simulate real-world document acquisition conditions. We use Qwen3-VL-2B and Qwen3-VL-4B as the foundation model for \method{}.
During the training of \method{} models, we use 4 NVIDIA A800 GPUs with DeepSpeed ZeRO Stage 2 for distributed full-parameter fine-tuning. The per-device batch size is set to 4, with gradient accumulation over 4 steps, resulting in an effective global batch size of 64. We train the models for 15K steps using a learning rate of 5e-6 and a cosine learning rate scheduler. The maximum sequence length is set to 8192, and the maximum image resolution is set to 1,605,632 pixels to support high-resolution document inputs. We use bfloat16 precision during training.

%% file: tables/overall_merged.tex
\begin{table*}[t]
\centering
\small
\begin{tabular}{l*{7}{c}c}
\toprule
\multicolumn{1}{l}{\multirow{2}{*}[-0.4ex]{\textbf{Method}}}
& \multicolumn{3}{c}{\textbf{Constrained-Category}}
& \multicolumn{4}{c}{\textbf{Open-Category}}
& \multicolumn{1}{c}{\multirow{2}{*}[-0.4ex]{\textbf{Avg.}}}
\\
\cmidrule(lr){2-4}\cmidrule(lr){5-8}
& Bus. Trans.
& Pub. Serv.
& Reg. Rec.
& Receipt
& Form
& Invoice
& Contract
&
\\
\midrule
\multicolumn{9}{l}{\textbf{On-Server LMMs}} \\
\midrule
Claude-Sonnet-4.5  & 64.53 & 68.29 & 66.33 & 39.62 & 40.21 & 36.65 & 42.38 & 51.14 \\
GPT-4o             & 68.16 & 67.14 & 73.14 & 59.09 & 39.45 & 46.20 & 42.01 & 56.46 \\
GPT-5              & 64.62 & 67.37 & 71.86 & 60.12 & 41.03 & 47.57 & 40.85 & 56.20 \\
Qwen-VL-Max        & 74.19 & 82.51 & 76.23 & 70.35 & 67.63 & 68.71 & 73.12 & 73.25 \\
Qwen3-VL-Plus      & 76.00 & 83.26 & 81.70 & 69.70 & 68.34 & 71.52 & 72.88 & 74.77 \\
Gemini-3-Pro       & 77.03 & 86.58 & 83.87 & 86.03 & 79.79 & 82.84 & 77.93 & 82.01 \\
\midrule
\multicolumn{9}{l}{\textbf{On-Device LMMs}} \\
\midrule
Ministral-3-8B     & 33.28 & 39.31 & 41.16 & 41.88 & 42.41 & 40.04 & 42.00 & 40.01 \\
MiniCPM-V4.5-8B    & 64.24 & 74.50 & 61.25 & 62.37 & 45.08 & 50.49 & 52.47 & 58.63 \\
InternVL3.5-8B     & 64.56 & 70.54 & 68.09 & 66.86 & 40.73 & 45.19 & 44.75 & 57.25 \\
MiMo-VL-7B-RL      & 67.91 & 74.11 & 63.22 & 64.48 & 54.72 & 54.78 & 68.22 & 63.92 \\
GLM-4.1V-9B        & 65.69 & 75.76 & 67.74 & 59.51 & 55.64 & 57.42 & 61.69 & 63.35 \\
Qwen3-VL-2B        & 64.03 & 68.00 & 72.98 & 59.14 & 48.77 & 50.86 & 58.57 & 60.34 \\
Qwen3-VL-4B        & 68.89 & 72.23 & 77.79 & 57.60 & 55.44 & 56.16 & 61.62 & 64.25 \\
\midrule
\multicolumn{9}{l}{\textbf{\method{}}} \\
\midrule
\method{}-2B         & 67.56 & 78.35 & 78.99 & 76.10 & 56.48 & 63.13 & 72.72 & 70.47 \\
\method{}-4B         & 70.38 & 80.73 & 80.97 & 78.15 & 59.97 & 64.89 & 75.36 & 72.92 \\
\bottomrule
\end{tabular}%
\caption{Overall Performance of \method{} Models and Baselines. \method{}-2B and \method{}-4B use Qwen3-VL-2B and Qwen3-VL-4B as their foundation models, respectively, and are fine-tuned for 15K steps on data synthesized by our \method{} framework.}
\label{tab:overall_merged}
\end{table*}

%% file: sections/result.tex
\section{Evaluation Results}
In this section, we train models using synthetic data generated by \method{} and evaluate their performance. We further examine the validity and scalability of the synthetic data.

\subsection{Overall Performance}
Table~\ref{tab:overall_merged} reports the overall KIE performance under both constrained- and open-category settings. Across both backbone sizes, fine-tuning on data synthesized by \method{} consistently improves the corresponding foundation models. This trend holds across document categories and evaluation settings, showing that the synthesized data provide broadly useful supervision for KIE rather than benefiting only a limited set of document types.

\method{} also achieves the strongest overall results among on-device LMMs. The 2B variant surpasses stronger baselines despite its smaller backbone, while \method{}-4B ranks first across all evaluated categories among on-device LMMs. Its performance is also close to that of strong proprietary on-server systems, indicating that improving training data can substantially narrow the capability gap between compact locally deployable models and much larger server-side LMMs. This result suggests that data quality and coverage can be as important as model scale for improving practical KIE performance under deployment constraints. The improvements are particularly pronounced in the open-category setting. Unlike constrained-category evaluation, this setting requires models to adapt to previously unseen document schemas and field definitions. The larger gains in this setting therefore suggest that \method{} does not simply improve extraction under known templates, but strengthens the model's ability to generalize to new document scenes. The improvements on challenging Receipt and Contract documents further support this conclusion, as these categories often involve diverse field organizations. Overall, the results show that scene-aware synthesis improves robustness of models to heterogeneous, schema-varying KIE scenarios.

\subsection{Scalability of Synthetic Data}

\input{figs/scaling_figures}

Figure~\ref{fig:data_scaling} further studies the scalability of \method{} by varying the amount of synthesized training data. For both 2B and 4B backbones, performance shows an overall upward trend as more generated data is introduced, indicating that the benefit of \method{} scales with data quantity.

The scaling behavior differs across evaluation settings. In the constrained-category setting, both models improve rapidly at early training stages and then enter a relatively stable range, suggesting that documents with fixed schemas and more regular field structures can be learned efficiently from a moderate amount of synthesized data. In contrast, the open-category setting continues to benefit from additional data, especially at later training stages. This is consistent with the greater diversity of open-category documents, where models must handle broader schema variations, layout patterns, and field organizations.
The average gain curve provides a more direct view of the contribution of synthesized data. Compared with the foundation model, both \method{}-2B and \method{}-4B obtain positive gains as the training data increases. The improvement is particularly strong for the 2B model, showing that smaller LMMs can extract substantial benefit from scalable synthetic supervision. Overall, these results demonstrate that \method{} is not only effective as a data augmentation method, but also exhibits favorable scaling behavior.

\subsection{Effectiveness Analysis}

\input{figs/error_analysis}

To understand how synthesized training data help models, we compare Qwen3-VL-4B with \method{}-4B using the field-level errors, as shown in Figure~\ref{fig:error_analysis}. We group field-level errors into seven types according to field names and document semantics: line-item fields, IDs/amounts, names/titles, dates/times, contact fields, contract clauses, and others. The last group contains heterogeneous residual fields and is omitted from the figure for clarity.

Training with \method{} reduces the total number of field-level errors by 17.8\%.
We further find that false positives and false negatives are reduced, indicating that the improvement is not caused by a conservative prediction strategy, but by better field localization and schema alignment. The largest reduction comes from line-item fields, including item names, quantities, prices, and subtotals. These fields often appear in dense tables with repeated rows, where Qwen3-VL-4B tends to miss entries or associate values with incorrect keys. 
\method{} also reduces errors on contract clauses and business identifiers, such as dispute resolution authority, receipt numbers, document numbers, and payment-related fields. These results suggest that scene-aware synthesis helps the model learn structural correspondences between layout regions and extraction schemas, especially in documents with repeated fields, business identifiers, and long-form contractual content. Overall, the primary benefit of synthesized data is not improved text recognition alone, but enhanced schema-aware extraction for documents with complex layouts.

%% file: figs/scaling_figures.tex
\begin{figure}[t]
    \centering
    \begin{minipage}[t]{0.495\columnwidth}
        \centering
        \includegraphics[width=\linewidth]{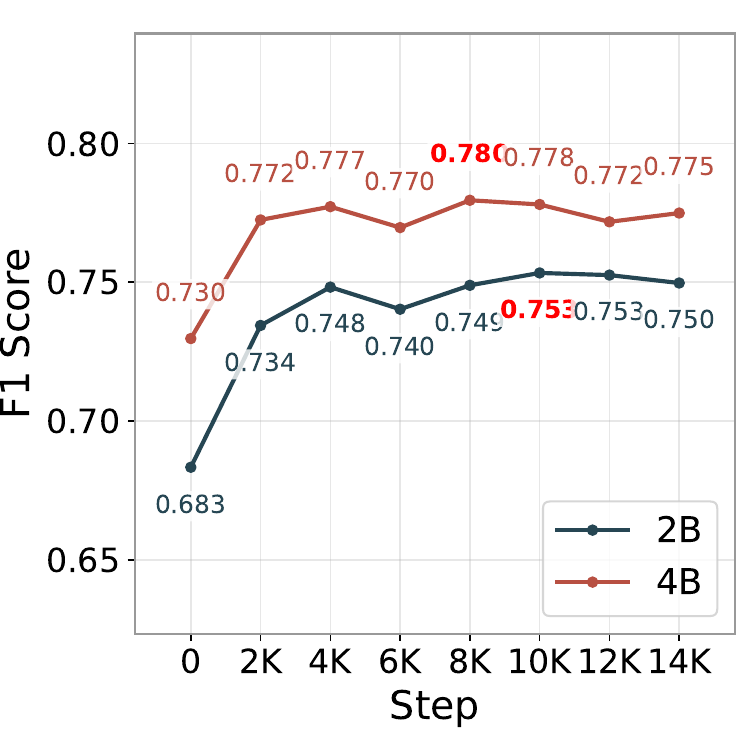}\\
        \footnotesize (a) Constrained KIE
    \end{minipage}
    \hspace{-0.3em}
    \begin{minipage}[t]{0.495\columnwidth}
        \centering
        \includegraphics[width=\linewidth]{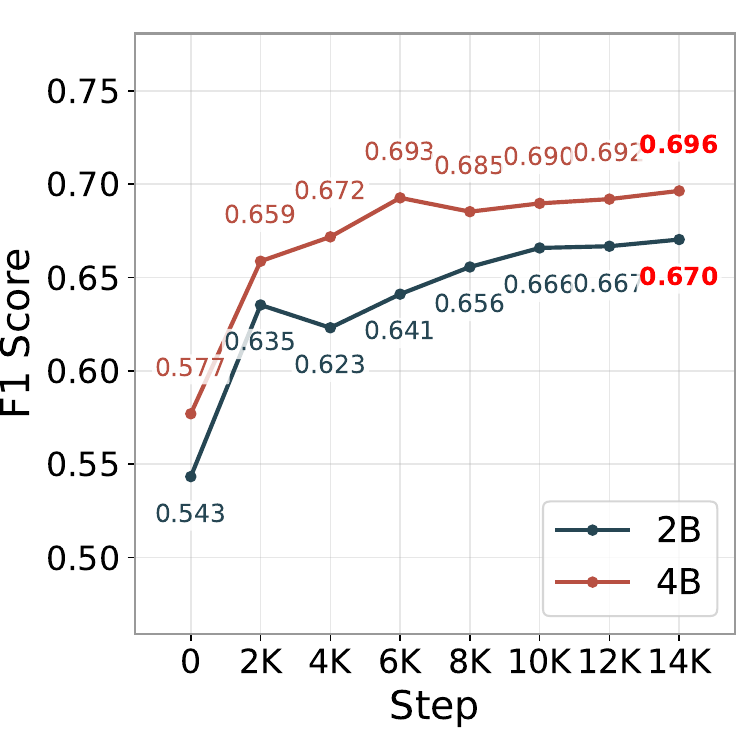}\\
        \footnotesize (b) Open KIE
    \end{minipage}
    \vspace{0.4em}

    \begin{minipage}[t]{0.495\columnwidth}
        \centering
        \includegraphics[width=\linewidth]{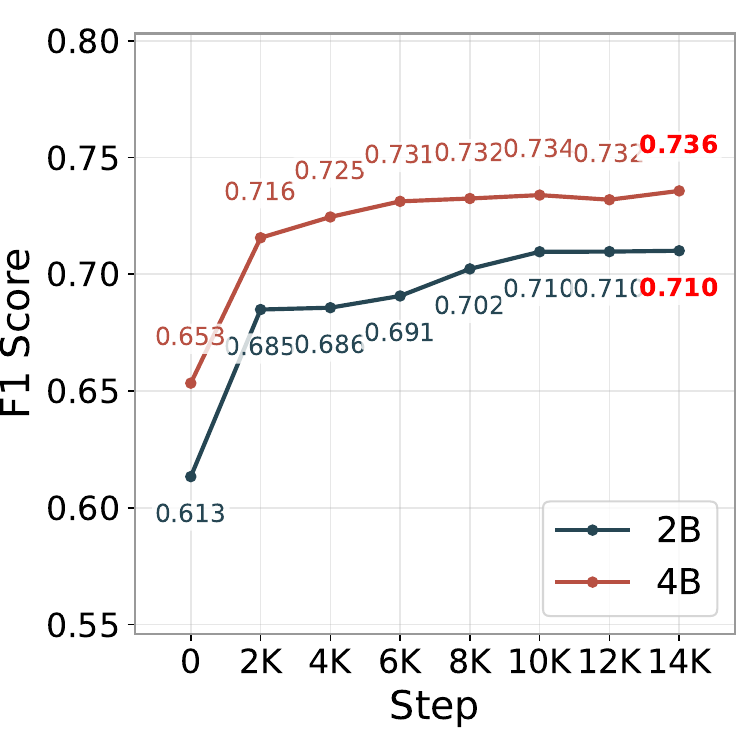}\\
        \footnotesize (c) Average F1
    \end{minipage}
    \hspace{-0.3em}
    \begin{minipage}[t]{0.495\columnwidth}
        \centering
        \includegraphics[width=\linewidth]{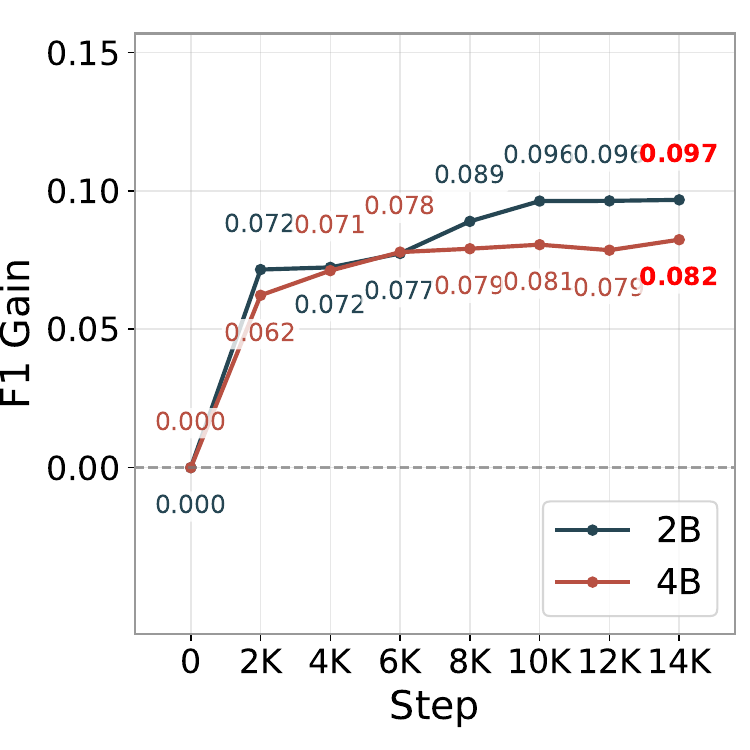}\\
        \footnotesize (d) Average F1 Gain
    \end{minipage}
    \caption{Performance Scaling with Synthetic Data Generated by \method{}. We report F1 scores on constrained- and open-category KIE, their average, and improvements over the foundation model.}
    \label{fig:data_scaling}
\end{figure}

%% file: figs/error_analysis.tex
\begin{figure}[t]
    \centering
    \includegraphics[width=\columnwidth]{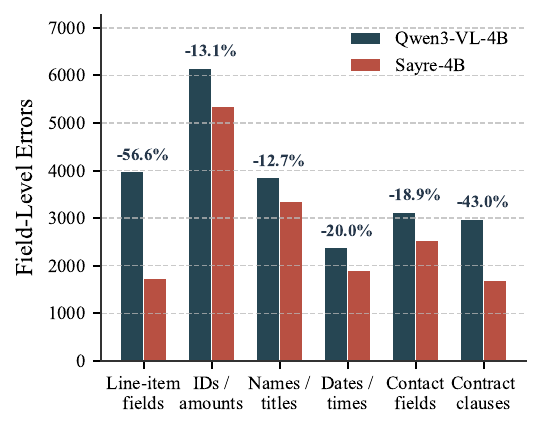}
    \caption{Field-Level Error Analysis of Qwen3-VL-4B and \method{}-4B across six interpretable error groups. Percentages above bars indicate the relative error reduction after training with synthesized data.}
    \label{fig:error_analysis}
\end{figure}

%% file: sections/conclusion.tex
\section{Conclusion}

We presented \method{}, a scene-aware document synthesis framework for improving compact LMMs on KIE. By generating diverse document--schema--annotation triples and expanding real-world failure cases into hard training examples, \method{} provides effective supervision for both constrained- and open-category extraction. Experiments show consistent gains over foundation models and strong on-device baselines, with particularly clear benefits for smaller backbones and open-category documents. These findings highlight scene-aware synthetic data as a practical path toward capable, locally deployable document understanding models.

\section*{Limitation}

This work is an initial exploration of using synthetic data to train Large Multimodal Models (LMMs) for Key Information Extraction (KIE), and several limitations remain. The current models still struggle with handwritten content, leading to degraded performance on mixed printed--handwritten documents. This limitation arises because \method{} cannot yet reliably synthesize realistic handwritten text and its associated visual variations.

%% file: custom.bib
@article{ji2026unikie,
  title={UNIKIE-BENCH: Benchmarking Large Multimodal Models for Key Information Extraction in Visual Documents},
  author={Ji, Yifan and Xu, Zhipeng and Liu, Zhenghao and Chen, Zulong and Zhang, Qian and Yang, Zhibo and Lin, Junyang and Gu, Yu and Yu, Ge and Sun, Maosong},
  journal={arXiv preprint arXiv:2602.07038},
  year={2026}
}

@inproceedings{simon2005viper,
  title={ViPER: augmenting automatic information extraction with visual perceptions},
  author={Simon, Kai and Lausen, Georg},
  booktitle={Proceedings of the 14th ACM international conference on Information and knowledge management},
  pages={381--388},
  year={2005}
}

@article{rombach2025deep,
  title={Deep learning based key information extraction from business documents: Systematic literature review},
  author={Rombach, Alexander Michael and Fettke, Peter},
  journal={ACM Computing Surveys},
  volume={58},
  number={2},
  pages={1--37},
  year={2025},
  publisher={ACM New York, NY}
}

@article{abdallah2024survey,
  title={A survey of recent approaches to form understanding in scanned documents},
  author={Abdallah, Abdelrahman and Eberharter, Daniel and Pfister, Zoe and Jatowt, Adam},
  journal={Artificial Intelligence Review},
  volume={57},
  number={12},
  pages={342},
  year={2024},
  publisher={Springer}
}

@article{sun2021spatial,
  title={Spatial dual-modality graph reasoning for key information extraction},
  author={Sun, Hongbin and Kuang, Zhanghui and Yue, Xiaoyu and Lin, Chenhao and Zhang, Wayne},
  journal={arXiv preprint arXiv:2103.14470},
  year={2021}
}

@inproceedings{katti2018chargrid,
  title={Chargrid: Towards understanding 2d documents},
  author={Katti, Anoop R and Reisswig, Christian and Guder, Cordula and Brarda, Sebastian and Bickel, Steffen and H{\"o}hne, Johannes and Faddoul, Jean Baptiste},
  booktitle={Proceedings of the 2018 Conference on Empirical Methods in Natural Language Processing},
  pages={4459--4469},
  year={2018}
}

@inproceedings{huang2019icdar,
  title={ICDAR2019 Competition on Scanned Receipt OCR and Information Extraction},
  author={Huang, Zheng and Chen, Kai and He, Jianhua and Bai, Xiang and
          Karatzas, Dimosthenis and Lu, Shijian and Jawahar, C. V.},
  booktitle={2019 International Conference on Document Analysis and Recognition},
  pages={1516--1520},
  year={2019},
  organization={IEEE}
}

@article{barboule2025survey,
  title={Survey on question answering over visually rich documents: Methods, challenges, and trends},
  author={Barboule, Camille and Piwowarski, Benjamin and Chabot, Yoan},
  journal={arXiv preprint arXiv:2501.02235},
  year={2025}
}

@inproceedings{huang2022layoutlmv3,
  title={Layoutlmv3: Pre-training for document ai with unified text and image masking},
  author={Huang, Yupan and Lv, Tengchao and Cui, Lei and Lu, Yutong and Wei, Furu},
  booktitle={Proceedings of the 30th ACM international conference on multimedia},
  pages={4083--4091},
  year={2022}
}

@inproceedings{bhattacharyya2025information,
  title={Information extraction from visually rich documents using LLM-based organization of documents into independent textual segments},
  author={Bhattacharyya, Aniket and Tripathi, Anurag and Das, Ujjal and Karmakar, Archan and Pathak, Amit and Gupta, Maneesh},
  booktitle={Proceedings of the 63rd Annual Meeting of the Association for Computational Linguistics (Volume 1: Long Papers)},
  pages={17241--17256},
  year={2025}
}

@article{ding2026deep,
  title={Deep learning based visually rich document content understanding: A survey},
  author={Ding, Yihao and Han, Soyeon Caren and Lee, Jean and Hovy, Eduard},
  journal={Artificial Intelligence Review},
  year={2026},
  publisher={Springer}
}

@article{xiong2026lang2act,
  title={Lang2Act: Fine-Grained Visual Reasoning through Self-Emergent Linguistic Toolchains},
  author={Xiong, Yuqi and Peng, Chunyi and Xu, Zhipeng and Liu, Zhenghao and Chen, Zulong and Yan, Yukun and Wang, Shuo and Gu, Yu and Yu, Ge},
  journal={arXiv preprint arXiv:2602.13235},
  year={2026}
}

@article{han2025mdocagent,
  title={Mdocagent: A multi-modal multi-agent framework for document understanding},
  author={Han, Siwei and Xia, Peng and Zhang, Ruiyi and Sun, Tong and Li, Yun and Zhu, Hongtu and Yao, Huaxiu},
  journal={arXiv preprint arXiv:2503.13964},
  year={2025}
}

@article{wang2025vrag,
  title={Vrag-rl: Empower vision-perception-based rag for visually rich information understanding via iterative reasoning with reinforcement learning},
  author={Wang, Qiuchen and Ding, Ruixue and Zeng, Yu and Chen, Zehui and Chen, Lin and Wang, Shihang and Xie, Pengjun and Huang, Fei and Zhao, Feng},
  journal={arXiv preprint arXiv:2505.22019},
  year={2025}
}

@inproceedings{li2022dit,
  title={Dit: Self-supervised pre-training for document image transformer},
  author={Li, Junlong and Xu, Yiheng and Lv, Tengchao and Cui, Lei and Zhang, Cha and Wei, Furu},
  booktitle={Proceedings of the 30th ACM international conference on multimedia},
  pages={3530--3539},
  year={2022}
}

@inproceedings{mao2024visually,
  title={Visually guided generative text-layout pre-training for document intelligence},
  author={Mao, Zhiming and Bai, Haoli and Hou, Lu and Shang, Lifeng and Jiang, Xin and Liu, Qun and Wong, Kam-Fai},
  booktitle={Proceedings of the 2024 Conference of the North American Chapter of the Association for Computational Linguistics: Human Language Technologies (Volume 1: Long Papers)},
  pages={4713--4730},
  year={2024}
}

@article{bai2025qwen2,
  title={Qwen2. 5-VL Technical Report},
  author={Bai, Shuai and Chen, Keqin and Liu, Xuejing and Wang, Jialin and Ge, Wenbin and Song, Sibo and Dang, Kai and Wang, Peng and Wang, Shijie and Tang, Jun and others},
  journal={arXiv preprint arXiv:2502.13923},
  year={2025}
}

@inproceedings{lu2025bounding,
  title={A Bounding Box is Worth One Token-Interleaving Layout and Text in a Large Language Model for Document Understanding},
  author={Lu, Jinghui and Yu, Haiyang and Wang, Yanjie and Ye, Yongjie and Tang, Jingqun and Yang, Ziwei and Wu, Binghong and Liu, Qi and Feng, Hao and Wang, Han and others},
  booktitle={Findings of the Association for Computational Linguistics: ACL 2025},
  pages={7252--7273},
  year={2025}
}

@inproceedings{zhang2024modeling,
  title={Modeling layout reading order as ordering relations for visually-rich document understanding},
  author={Zhang, Chong and Tu, Yi and Zhao, Yixi and Yuan, Chenshu and Chen, Huan and Zhang, Yue and Chai, Mingxu and Guo, Ya and Zhu, Huijia and Zhang, Qi and others},
  booktitle={Proceedings of the 2024 conference on empirical methods in natural language processing},
  pages={9658--9678},
  year={2024}
}

@inproceedings{zmigrod2024value,
  title={“What is the value of templates?” Rethinking Document Information Extraction Datasets for LLMs},
  author={Zmigrod, Ran and Shetty, Pranav and Sibue, Mathieu and Ma, Zhiqiang and Nourbakhsh, Armineh and Liu, Xiaomo and Veloso, Manuela},
  booktitle={Findings of the Association for Computational Linguistics: EMNLP 2024},
  pages={13162--13185},
  year={2024}
}

@inproceedings{jiang2025relayout,
  title={Relayout: Towards real-world document understanding via layout-enhanced pre-training},
  author={Jiang, Zhouqiang and Wang, Bowen and Chen, Junhao and Nakashima, Yuta},
  booktitle={Proceedings of the 31st International Conference on Computational Linguistics},
  pages={3778--3793},
  year={2025}
}

@article{liu2024focus,
  title={Focus anywhere for fine-grained multi-page document understanding},
  author={Liu, Chenglong and Wei, Haoran and Chen, Jinyue and Kong, Lingyu and Ge, Zheng and Zhu, Zining and Zhao, Liang and Sun, Jianjian and Han, Chunrui and Zhang, Xiangyu},
  journal={arXiv preprint arXiv:2405.14295},
  year={2024}
}

@inproceedings{zhang2025dockylin,
  title={Dockylin: A large multimodal model for visual document understanding with efficient visual slimming},
  author={Zhang, Jiaxin and Yang, Wentao and Lai, Songxuan and Xie, Zecheng and Jin, Lianwen},
  booktitle={Proceedings of the AAAI Conference on Artificial Intelligence},
  volume={39},
  number={9},
  pages={9923--9932},
  year={2025}
}

@inproceedings{li2024monkey,
  title={Monkey: Image resolution and text label are important things for large multi-modal models},
  author={Li, Zhang and Yang, Biao and Liu, Qiang and Ma, Zhiyin and Zhang, Shuo and Yang, Jingxu and Sun, Yabo and Liu, Yuliang and Bai, Xiang},
  booktitle={proceedings of the IEEE/CVF conference on computer vision and pattern recognition},
  pages={26763--26773},
  year={2024}
}

@article{xie2024wukong,
  title={Wukong: A large multimodal model for efficient long pdf reading with end-to-end sparse sampling},
  author={Xie, Xudong and Yan, Hao and Yin, Liang and Liu, Yang and Ding, Jing and Liao, Minghui and Liu, Yuliang and Chen, Wei and Bai, Xiang},
  journal={arXiv preprint arXiv:2410.05970},
  year={2024}
}

@article{zhang2024document,
  title={Document parsing unveiled: Techniques, challenges, and prospects for structured information extraction},
  author={Zhang, Qintong and Wang, Bin and Huang, Victor Shea-Jay and Zhang, Junyuan and Wang, Zhengren and Liang, Hao and He, Conghui and Zhang, Wentao},
  journal={arXiv preprint arXiv:2410.21169},
  year={2024}
}

@article{ke2025large,
  title={Large language models in document intelligence: A comprehensive survey, recent advances, challenges, and future trends},
  author={Ke, Wenjun and Zheng, Yifan and Li, Yining and Xu, Hengyuan and Nie, Dong and Wang, Peng and He, Yao},
  journal={ACM Transactions on Information Systems},
  volume={44},
  number={1},
  pages={1--64},
  year={2025},
  publisher={ACM New York, NY}
}

@inproceedings{hu2024mplug,
  title={mplug-docowl 1.5: Unified structure learning for ocr-free document understanding},
  author={Hu, Anwen and Xu, Haiyang and Ye, Jiabo and Yan, Ming and Zhang, Liang and Zhang, Bo and Zhang, Ji and Jin, Qin and Huang, Fei and Zhou, Jingren},
  booktitle={Findings of the Association for Computational Linguistics: EMNLP 2024},
  pages={3096--3120},
  year={2024}
}

@article{liu2026textmonkey,
  title={Textmonkey: An ocr-free large multimodal model for understanding document},
  author={Liu, Yuliang and Yang, Biao and Liu, Qiang and Li, Zhang and Ma, Zhiyin and Zhang, Shuo and Bai, Xiang},
  journal={IEEE Transactions on Pattern Analysis and Machine Intelligence},
  year={2026},
  publisher={IEEE}
}

@article{liu2026see,
  title={See then tell: Enhancing key information extraction with vision grounding},
  author={Liu, Shuhang and Zhang, Zhenrong and Hu, Pengfei and Ma, Jiefeng and Du, Jun and Wang, Qing and Zhang, Jianshu and Liu, Chenyu},
  journal={Neurocomputing},
  pages={132858},
  year={2026},
  publisher={Elsevier}
}

@inproceedings{cao2023genkie,
  title={Genkie: Robust generative multimodal document key information extraction},
  author={Cao, Panfeng and Wang, Ye and Zhang, Qiang and Meng, Zaiqiao},
  booktitle={Findings of the Association for Computational Linguistics: EMNLP 2023},
  pages={14702--14713},
  year={2023}
}

@article{shen2022vila,
  title={VILA: Improving structured content extraction from scientific PDFs using visual layout groups},
  author={Shen, Zejiang and Lo, Kyle and Wang, Lucy Lu and Kuehl, Bailey and Weld, Daniel S and Downey, Doug},
  journal={Transactions of the Association for Computational Linguistics},
  volume={10},
  pages={376--392},
  year={2022},
  publisher={MIT Press One Broadway, 12th Floor, Cambridge, Massachusetts 02142, USA~…}
}

@article{shen2026ocr,
  title={OCR or Not? Rethinking Document Information Extraction in the MLLMs Era with Real-World Large-Scale Datasets},
  author={Shen, Jiyuan and Yuan, Peiyue and Ghosh, Atin and Mai, Yifan and Dahlmeier, Daniel},
  journal={arXiv preprint arXiv:2603.02789},
  year={2026}
}

@article{xu2021layoutxlm,
  title={Layoutxlm: Multimodal pre-training for multilingual visually-rich document understanding},
  author={Xu, Yiheng and Lv, Tengchao and Cui, Lei and Wang, Guoxin and Lu, Yijuan and Florencio, Dinei and Zhang, Cha and Wei, Furu},
  journal={arXiv preprint arXiv:2104.08836},
  year={2021}
}

@inproceedings{shim2025revise,
  title={Revise: A framework for revising ocred text in practical information systems with data contamination strategy},
  author={Shim, Gyuho and Hong, Seongtae and Lim, Heui-Seok},
  booktitle={Proceedings of the 63rd Annual Meeting of the Association for Computational Linguistics (Volume 6: Industry Track)},
  pages={1423--1434},
  year={2025}
}

@inproceedings{agarwal2025fs,
  title={FS-DAG: Few shot domain adapting graph networks for visually rich document understanding},
  author={Agarwal, Amit and Panda, Srikant and Pachauri, Kulbhushan},
  booktitle={Proceedings of the 31st International Conference on Computational Linguistics: Industry Track},
  pages={100--114},
  year={2025}
}

@inproceedings{li2021selfdoc,
  title={Selfdoc: Self-supervised document representation learning},
  author={Li, Peizhao and Gu, Jiuxiang and Kuen, Jason and Morariu, Vlad I and Zhao, Handong and Jain, Rajiv and Manjunatha, Varun and Liu, Hongfu},
  booktitle={Proceedings of the IEEE/CVF conference on computer vision and pattern recognition},
  pages={5652--5660},
  year={2021}
}

@inproceedings{peng2022ernie,
  title={Ernie-layout: Layout knowledge enhanced pre-training for visually-rich document understanding},
  author={Peng, Qiming and Pan, Yinxu and Wang, Wenjin and Luo, Bin and Zhang, Zhenyu and Huang, Zhengjie and Cao, Yuhui and Yin, Weichong and Chen, Yongfeng and Zhang, Yin and others},
  booktitle={Findings of the Association for Computational Linguistics: EMNLP 2022},
  pages={3744--3756},
  year={2022}
}

@inproceedings{wang2022lilt,
  title={Lilt: A simple yet effective language-independent layout transformer for structured document understanding},
  author={Wang, Jiapeng and Jin, Lianwen and Ding, Kai},
  booktitle={Proceedings of the 60th Annual Meeting of the Association for Computational Linguistics (Volume 1: Long Papers)},
  pages={7747--7757},
  year={2022}
}

@inproceedings{wang2024docllm,
  title={Docllm: A layout-aware generative language model for multimodal document understanding},
  author={Wang, Dongsheng and Raman, Natraj and Sibue, Mathieu and Ma, Zhiqiang and Babkin, Petr and Kaur, Simerjot and Pei, Yulong and Nourbakhsh, Armineh and Liu, Xiaomo},
  booktitle={Proceedings of the 62nd Annual Meeting of the Association for Computational Linguistics (Volume 1: Long Papers)},
  pages={8529--8548},
  year={2024}
}

@inproceedings{li2021structext,
  title={Structext: Structured text understanding with multi-modal transformers},
  author={Li, Yulin and Qian, Yuxi and Yu, Yuechen and Qin, Xiameng and Zhang, Chengquan and Liu, Yan and Yao, Kun and Han, Junyu and Liu, Jingtuo and Ding, Errui},
  booktitle={Proceedings of the 29th ACM international conference on multimedia},
  pages={1912--1920},
  year={2021}
}

@inproceedings{appalaraju2021docformer,
  title={Docformer: End-to-end transformer for document understanding},
  author={Appalaraju, Srikar and Jasani, Bhavan and Kota, Bhargava Urala and Xie, Yusheng and Manmatha, R},
  booktitle={Proceedings of the IEEE/CVF international conference on computer vision},
  pages={993--1003},
  year={2021}
}

@inproceedings{hong2022bros,
  title={Bros: A pre-trained language model focusing on text and layout for better key information extraction from documents},
  author={Hong, Teakgyu and Kim, Donghyun and Ji, Mingi and Hwang, Wonseok and Nam, Daehyun and Park, Sungrae},
  booktitle={Proceedings of the AAAI Conference on Artificial Intelligence},
  volume={36},
  number={10},
  pages={10767--10775},
  year={2022}
}

@inproceedings{skalicky2022business,
  title={Business document information extraction: Towards practical benchmarks},
  author={Skalick{\`y}, Maty{\'a}{\v{s}} and {\v{S}}imsa, {\v{S}}t{\v{e}}p{\'a}n and U{\v{r}}i{\v{c}}{\'a}{\v{r}}, Michal and {\v{S}}ulc, Milan},
  booktitle={International Conference of the Cross-Language Evaluation Forum for European Languages},
  pages={105--117},
  year={2022},
  organization={Springer}
}

@inproceedings{zhang2020trie,
  title={TRIE: end-to-end text reading and information extraction for document understanding},
  author={Zhang, Peng and Xu, Yunlu and Cheng, Zhanzhan and Pu, Shiliang and Lu, Jing and Qiao, Liang and Niu, Yi and Wu, Fei},
  booktitle={Proceedings of the 28th ACM International Conference on Multimedia},
  pages={1413--1422},
  year={2020}
}

@inproceedings{zhang2023reading,
  title={Reading order matters: information extraction from visually-rich documents by token path prediction},
  author={Zhang, Chong and Guo, Ya and Tu, Yi and Chen, Huan and Tang, Jinyang and Zhu, Huijia and Zhang, Qi and Gui, Tao},
  booktitle={Proceedings of the 2023 Conference on Empirical Methods in Natural Language Processing},
  pages={13716--13730},
  year={2023}
}

@inproceedings{hwang2021cost,
  title={Cost-effective end-to-end information extraction for semi-structured document images},
  author={Hwang, Wonseok and Lee, Hyunji and Yim, Jinyeong and Kim, Geewook and Seo, Minjoon},
  booktitle={Proceedings of the 2021 Conference on Empirical Methods in Natural Language Processing},
  pages={3375--3383},
  year={2021}
}

@inproceedings{palm2017cloudscan,
  title={Cloudscan-a configuration-free invoice analysis system using recurrent neural networks},
  author={Palm, Rasmus Berg and Winther, Ole and Laws, Florian},
  booktitle={2017 14th IAPR International Conference on Document Analysis and Recognition (ICDAR)},
  volume={1},
  pages={406--413},
  year={2017},
  organization={IEEE}
}

@article{cui2021document,
  title={Document ai: Benchmarks, models and applications},
  author={Cui, Lei and Xu, Yiheng and Lv, Tengchao and Wei, Furu},
  journal={arXiv preprint arXiv:2111.08609},
  year={2021}
}

@inproceedings{chiticariu2010enterprise,
  title={Enterprise information extraction: recent developments and open challenges},
  author={Chiticariu, Laura and Li, Yunyao and Raghavan, Sriram and Reiss, Frederick R},
  booktitle={Proceedings of the 2010 ACM SIGMOD International Conference on Management of data},
  pages={1257--1258},
  year={2010}
}

@inproceedings{yang2025cc,
  title={Cc-ocr: A comprehensive and challenging ocr benchmark for evaluating large multimodal models in literacy},
  author={Yang, Zhibo and Tang, Jun and Li, Zhaohai and Wang, Pengfei and Wan, Jianqiang and Zhong, Humen and Liu, Xuejing and Yang, Mingkun and Wang, Peng and Bai, Shuai and others},
  booktitle={Proceedings of the IEEE/CVF International Conference on Computer Vision},
  pages={21744--21754},
  year={2025}
}

@article{aumann2006visual,
  title={Visual information extraction},
  author={Aumann, Yonatan and Feldman, Ronen and Liberzon, Yair and Rosenfeld, Benjamin and Schler, Jonathan},
  journal={Knowledge and Information Systems},
  volume={10},
  number={1},
  pages={1--15},
  year={2006},
  publisher={Springer}
}

@incollection{grishman1997information,
  title={Information extraction: Techniques and challenges},
  author={Grishman, Ralph},
  booktitle={International summer school on information extraction},
  pages={10--27},
  year={1997},
  publisher={Springer}
}

@article{ge2024scaling,
  title={Scaling synthetic data creation with 1,000,000,000 personas},
  author={Ge, Tao and Chan, Xin and Wang, Xiaoyang and Yu, Dian and Mi, Haitao and Yu, Dong},
  journal={arXiv preprint arXiv:2406.20094},
  year={2024}
}

@article{liu2026ministral,
  title={Ministral 3},
  author={Liu, Alexander H and Khandelwal, Kartik and Subramanian, Sandeep and Jouault, Victor and Rastogi, Abhinav and Sad{\'e}, Adrien and Jeffares, Alan and Jiang, Albert and Cahill, Alexandre and Gavaudan, Alexandre and others},
  journal={arXiv preprint arXiv:2601.08584},
  year={2026}
}

@article{yu2025minicpm,
  title={Minicpm-v 4.5: Cooking efficient mllms via architecture, data, and training recipe},
  author={Yu, Tianyu and Wang, Zefan and Wang, Chongyi and Huang, Fuwei and Ma, Wenshuo and He, Zhihui and Cai, Tianchi and Chen, Weize and Huang, Yuxiang and Zhao, Yuanqian and others},
  journal={arXiv preprint arXiv:2509.18154},
  year={2025}
}

@article{bai2025qwen3,
  title={Qwen3-vl technical report},
  author={Bai, Shuai and Cai, Yuxuan and Chen, Ruizhe and Chen, Keqin and Chen, Xionghui and Cheng, Zesen and Deng, Lianghao and Ding, Wei and Gao, Chang and Ge, Chunjiang and others},
  journal={arXiv preprint arXiv:2511.21631},
  year={2025}
}

@article{wang2025internvl3,
  title={Internvl3. 5: Advancing open-source multimodal models in versatility, reasoning, and efficiency},
  author={Wang, Weiyun and Gao, Zhangwei and Gu, Lixin and Pu, Hengjun and Cui, Long and Wei, Xingguang and Liu, Zhaoyang and Jing, Linglin and Ye, Shenglong and Shao, Jie and others},
  journal={arXiv preprint arXiv:2508.18265},
  year={2025}
}

@inproceedings{stanislawek2021kleister,
  title={Kleister: key information extraction datasets involving long documents with complex layouts},
  author={Stanis{\l}awek, Tomasz and Grali{\'n}ski, Filip and Wr{\'o}blewska, Anna and Lipi{\'n}ski, Dawid and Kaliska, Agnieszka and Rosalska, Paulina and Topolski, Bartosz and Biecek, Przemys{\l}aw},
  booktitle={International Conference on Document Analysis and Recognition},
  pages={564--579},
  year={2021},
  organization={Springer}
}

@inproceedings{hwang2019post,
  title={Post-OCR parsing: building simple and robust parser via BIO tagging},
  author={Hwang, Wonseok and Kim, Seonghyeon and Seo, Minjoon and Yim, Jinyeong and Park, Seunghyun and Park, Sungrae and Lee, Junyeop and Lee, Bado and Lee, Hwalsuk},
  booktitle={Workshop on Document Intelligence at NeurIPS 2019},
  year={2019}
}

@inproceedings{xu2020layoutlm,
  title={Layoutlm: Pre-training of text and layout for document image understanding},
  author={Xu, Yiheng and Li, Minghao and Cui, Lei and Huang, Shaohan and Wei, Furu and Zhou, Ming},
  booktitle={Proceedings of the 26th ACM SIGKDD international conference on knowledge discovery \& data mining},
  pages={1192--1200},
  year={2020}
}

@article{poznanski2025olmocr,
  title={olmocr 2: Unit test rewards for document ocr},
  author={Poznanski, Jake and Soldaini, Luca and Lo, Kyle},
  journal={arXiv preprint arXiv:2510.19817},
  year={2025}
}

@inproceedings{guo2018cloud,
  title     = {Cloud-Based or On-Device: An Empirical Study of Mobile Deep Inference},
  author    = {Guo, Tian},
  booktitle = {2018 IEEE International Conference on Cloud Engineering},
  pages     = {184--190},
  year      = {2018},
  publisher = {IEEE Computer Society}
}

@inproceedings{kang2017neurosurgeon,
  title     = {Neurosurgeon: Collaborative Intelligence Between the Cloud and Mobile Edge},
  author    = {Kang, Yiping and Hauswald, Johann and Gao, Cao and Rovinski, Austin and
               Mudge, Trevor N. and Mars, Jason and Tang, Lingjia},
  booktitle = {Proceedings of the Twenty-Second International Conference on
               Architectural Support for Programming Languages and Operating Systems},
  pages     = {615--629},
  year      = {2017},
  publisher = {ACM}
}

@article{bensch2021kie,
  title   = {Key Information Extraction From Documents: Evaluation And Generator},
  author  = {Bensch, Oliver and Popa, Mirela and Spille, Constantin},
  journal = {arXiv preprint arXiv:2106.14624},
  year    = {2021}
}

@article{simsa2023docile,
  title   = {{DocILE} Benchmark for Document Information Localization and Extraction},
  author  = {{\v{S}}imsa, {\v{S}}t{\v{e}}p{\'a}n and
             {\v{S}}ulc, Milan and
             U{\v{r}}i{\v{c}}{\'a}{\v{r}}, Michal and
             Patel, Yash and
             Hamdi, Ahmed and
             Koci{\'a}n, Mat{\v{e}}j and
             Skalick{\`y}, Maty{\'a}{\v{s}} and
             Matas, Ji{\v{r}}{\'\i} and
             Doucet, Antoine and
             Coustaty, Micka{\"e}l and
             Karatzas, Dimosthenis},
  journal = {arXiv preprint arXiv:2302.05658},
  year    = {2023}
}

@inproceedings{dua2025flexdoc,
  title     = {{FlexDoc}: Parameterized Sampling for Diverse Multilingual
               Synthetic Documents for Training Document Understanding Models},
  author    = {Dua, Karan and Patel, Hitesh Laxmichand and Mittal, Puneet and
               Gupta, Ranjeet and Agarwal, Amit and Pabolu, Praneet and
               Panda, Srikant and Meghwani, Hansa and Horwood, Graham and
               Shah, Fahad},
  booktitle = {Proceedings of the 2025 Conference on Empirical Methods in
               Natural Language Processing: Industry Track},
  pages     = {1500--1521},
  year      = {2025},
  address   = {Suzhou, China},
  publisher = {Association for Computational Linguistics}
}

@article{team2025mimo,
  title={MiMo: Unlocking the Reasoning Potential of Language Model -- From Pretraining to Posttraining},
  author={Xia, Bingquan and Shen, Bowen and Zhu, Dawei and Zhang, Di and Wang, Gang and Zhang, Hailin and Liu, Huaqiu and Xiao, Jiebao and Dong, Jinhao and Zhao, Liang and Li, Peidian and Wang, Peng and Yu, Shihua and Chen, Shimao and Wang, Weikun and Ma, Wenhan and Deng, Xiangwei and Huang, Yi and Song, Yifan and Jiang, Zihan and others},
  journal={arXiv preprint arXiv:2505.07608},
  year={2025}
}

@article{team2025glm,
  title={ChatGLM: A Family of Large Language Models from {GLM}-130{B} to {GLM}-4 All Tools},
  author={Zeng, Aohan and Xu, Bin and Wang, Bowen and Zhang, Chenhui and Yin, Da and Zhang, Dan and Rojas, Diego and Feng, Guanyu and Zhao, Hanlin and Lai, Hanyu and Yu, Hao and Wang, Hongning and Sun, Jiadai and Zhang, Jiajie and Cheng, Jiale and Gui, Jiayi and Tang, Jie and Zhang, Jing and Sun, Jingyu and Li, Juanzi and others},
  journal={arXiv preprint arXiv:2406.12793},
  year={2024}
}
